# An Open-Source User-Friendly Interface for Simulating Magnetic Soft Robots using Simulation Open Framework Architecture (SOFA)

Carla Wehner[a]*, Finn Schubert[b]*, Heiko Hellkamp[b]*, Julius Hahnewald[b]*, Kilian Schäfer[c], Muhammad Bilal Khan[c‡], Oliver Gutfleisch[c]

[a]Department of Human Science, Technical University Darmstadt, 64283 Darmstadt, Germany.

[b]Department of Computer Science, Technical University Darmstadt, 64289 Darmstadt, Germany.

[c]Functional Materials, Institute of Materials Science, Technical University Darmstadt, 64287 Darmstadt, Germany.

[‡]Correspondence: Muhammad.khan3@tu-darmstadt.de

**Abstract**

Soft robots, particularly magnetic soft robots, require specialized simulation tools to accurately model their deformation under external magnetic fields. However, existing platforms often lack dedicated support for magnetic materials, making them difficult to use for researchers at different expertise levels. This work introduces an open-source, user-friendly simulation interface using the simulation open framework architecture (SOFA), specifically designed to model magnetic soft robots. The tool enables users to define material properties, apply magnetic fields, and observe resulting deformations in real time. By integrating intuitive controls and stress analysis capabilities, it aims to bridge the gap between theoretical modeling and practical design. Four benchmark models—a beam, three and four-finger grippers, and a butterfly—demonstrate its functionality. The software's ease of use makes it accessible to both beginners and advanced researchers. Future improvements will refine accuracy through experimental validation and comparison with industry-standard finite element solvers, ensuring realistic and predictive simulations of magnetic soft robots.

## 1. Introduction

Soft robotics is revolutionizing the field of robotic design by utilizing flexible, compliant materials that enable robots to bend, stretch, and deform dynamically [1]. Unlike traditional rigid robots, soft robots leverage material elasticity to achieve motion, making them particularly effective for biomedical applications, industrial automation, and bio-inspired systems [1-2]. Among various actuation methods, magnetic soft robots offer unique advantages due to their remote controllability, rapid response, and adaptability in confined environments [3-6]. However, accurately modeling their behavior remains a significant challenge due to the complex interplay between magnetic fields, elasticity, and nonlinear deformations [3-4]. Existing simulation tools lack specialized features for modeling magnetically actuated materials in an accessible and user-friendly manner, limiting their adoption among researchers and engineers.

---

*Development team, equal contributions (names in alphabetical order)

Magnetically actuated soft robots incorporate magnetic particles within a flexible polymer matrix, allowing controlled deformations under an external magnetic field [3]. This enables remote manipulation, making it particularly useful in applications such as minimally invasive surgery, targeted drug delivery, and robotic grippers [1-5]. Despite these advantages, modeling magnetic soft robots is inherently complex, requiring the simulation of coupled magneto-mechanical interactions and nonlinear material behaviors. Traditional finite element method (FEM) solvers provide accurate simulations, but often require extensive expertise, computational resources, and setup time [7]. Additionally, existing platforms lack real-time interaction capabilities, limiting their effectiveness for iterative design and control development [7]. This work presents an open-source simulation tool designed to bridge these gaps by integrating magnetic soft robot modeling using the SOFA framework [8]. The tool provides an intuitive graphical user interface (GUI) that allows users to define material properties, define magnetic fields, and visualize real-time deformations. By enabling direct parameter adjustments and including built-in stress analysis, the software facilitates rapid design exploration and optimization. This modular and extensible program is aimed to be accessible to users at all expertise levels, supporting research in soft robotics and practical engineering applications.

## 2. Results and Discussion

The simulation environment consists of three core components: a GUI for user interaction, a physics engine leveraging SOFA's FEM capabilities for real-time deformation calculations [7-8], and a visualization module that provides stress analysis and feedback. The software architecture is designed for modularity, ensuring compatibility with CAD tools and allowing for future expansions. The software is open source and hosted in a GitHub repository with all installation and run instructions [9].

In terms of software architecture, computational efficiency is granted by SOFA and its fast C++ implementation, whereas the modular architecture makes the software easy to maintain and extensible for future functionalities. The GUI enables users to import CAD-generated soft robot models, set material and actuation parameters, and visualize results interactively (Figure 1). SOFA computes magnetically induced deformations based on FEM formulations, providing real-time feedback for design optimization. The GUI simplifies the simulation process by allowing users to import soft robotic models in standard formats such as STL (separately processed to be MSH mesh) of the same imported object. It enables intuitive adjustments to material and magnetic properties, including young's modulus, poisson's ratio, density, and remanence, and actuation parameters such as magnetic field strength and its orientation (Figure 2(A-C)). The SOFA-based physics engine computes soft robot deformations under external magnetic fields in real time. The integrated stress analysis module evaluates strain distributions and mechanical integrity, helping users optimize material properties and actuation strategies [9]. The ability to adjust simulation parameters interactively allows for rapid iterations in design refinement.

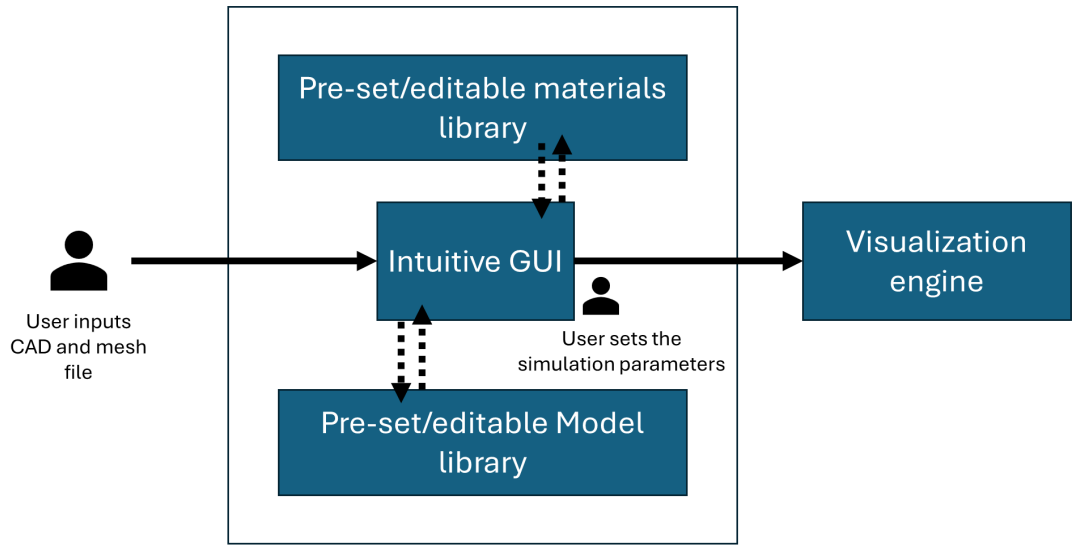

**Figure 1.** An overview of the developed user interface. A user imports a new model in STL format and a MSH mesh file, or selects a model from the model library. Then by assigning different required simulation parameters, the user can visualize the animated output in SOFA.

To demonstrate its capabilities, the software includes four predefined magnetic soft robotic models:

1. Beam Model – A simple structure used to test bending behavior (Figure 2(D-E)).

2. Three-Finger Gripper – A robotic gripper actuated via magnetic forces (Figure 3(A-B)).

3. Four-Finger Gripper – A more complex gripping mechanism with enhanced dexterity (Figure 3(C-D)).

4. Butterfly Model – A bio-inspired structure with shape-morphing capabilities, inspired by recent work in 3D-printed magnetic soft robots [5-6] (Figure 3(E-F)).

These examples show the versatility of the tool in handling different actuation scenarios and soft robotic applications.

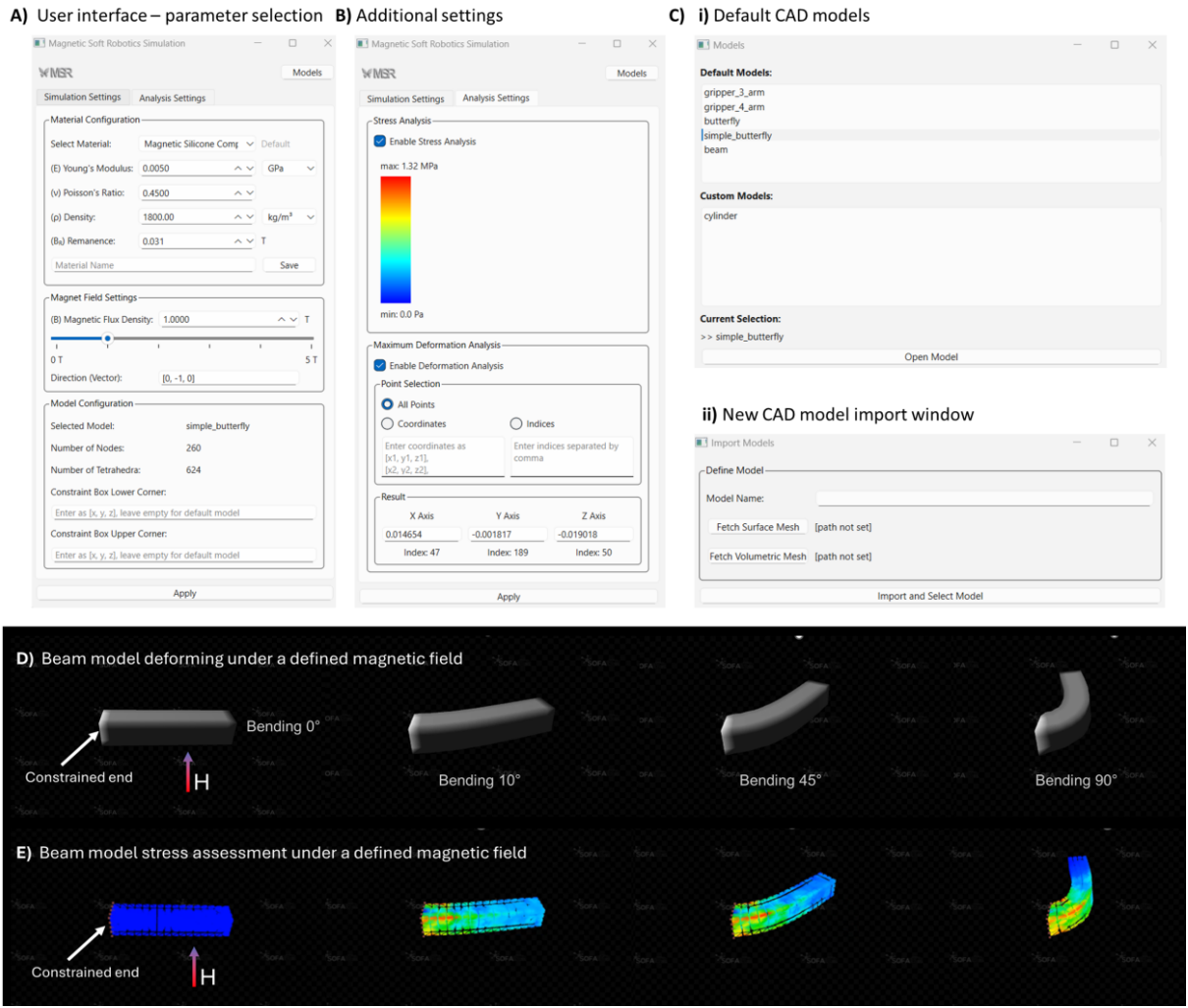

**Figure 2.** A) A detailed view of different simulation parameters in the GUI that can be set by a user. B) Additional simulation settings allowing a user to obtain stress estimate on the CAD model under a defined magnetic field, in addition to the deformation behavior. C) Model choice window showing the possibility of using i) pre-defined models, ii) importing new CAD models. D-E) shows a beam model deforming under a defined magnetic field and stress behavior during the same deformation.

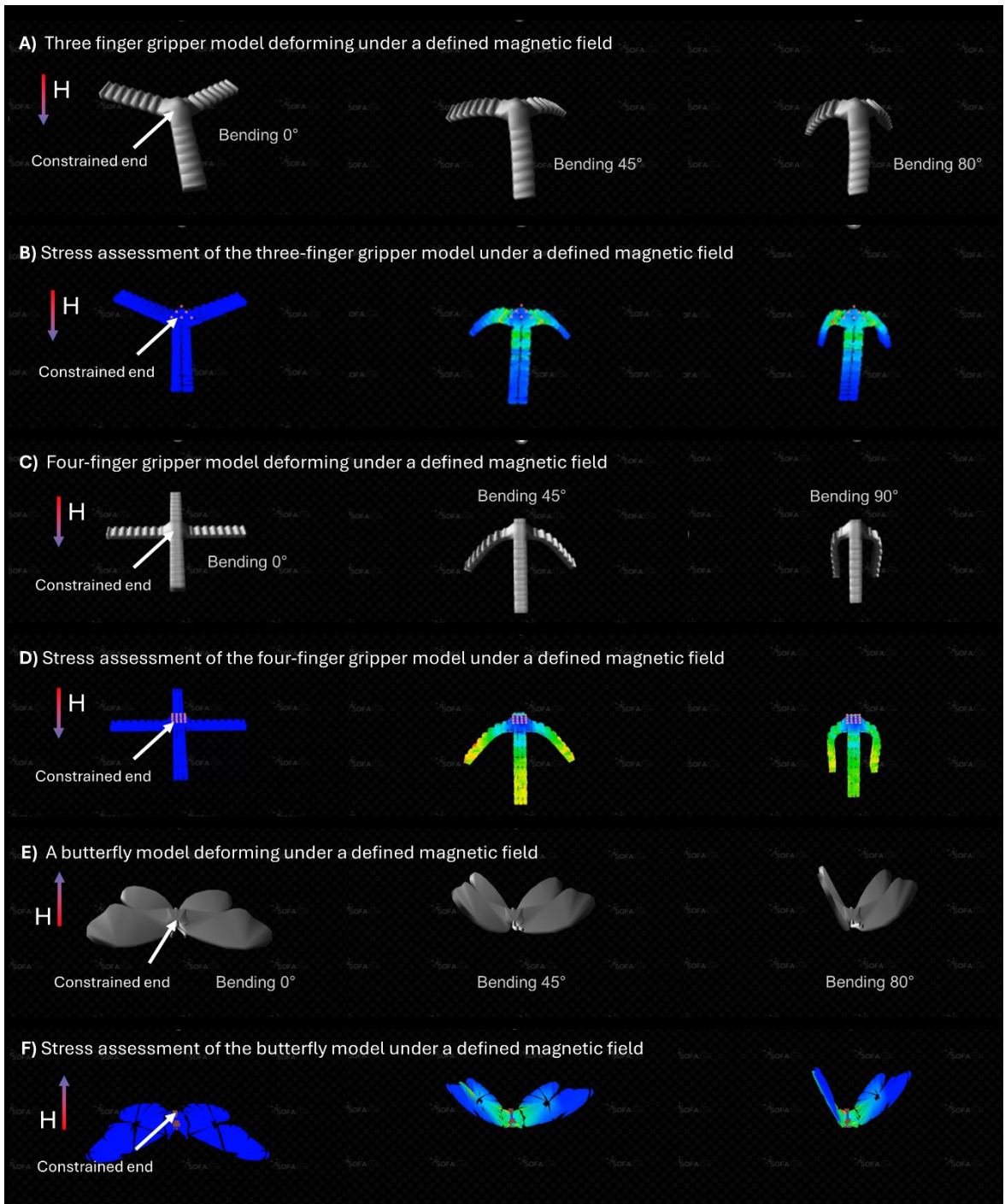

**Figure 3.** Examples of soft robotic models in the software deforming under applied magnetic fields with corresponding stress assessment. A-B) shows a three-finger gripper model deforming under a defined magnetic field. C-D) shows a four-finger gripper model deforming under a defined magnetic field. E-F) shows a butterfly model deforming under a defined magnetic field.

## 5. Conclusion and Outlook

This work introduces an open-source, real-time simulation tool for modeling magnetic soft robots within the SOFA framework. By combining user-friendly controls, real-time deformation simulations, and built-in stress analysis, the tool provides researchers and engineers with a useful resource for soft robotics development. Future validation efforts will focus on comparing software-generated deformations with real-world experiments. To assess the tool's accuracy, simulation results will be compared with established FEM solvers, such as COMSOL Multiphysics, ANSYS Mechanical, and Abaqus [10]. These comparisons will ensure that the simulation tool meets established standards for predictive reliability and mechanical analysis. Other planned improvements include enhanced magnetic field modeling to support spatially varying fields, GPU acceleration to improve computation speed, and the introduction of a scripting interface for advanced users. By bridging the gap between simulation and real-world performance, this tool has the potential to accelerate the design and deployment of magnetic soft robotic systems.

## Acknowledgement

This work was supported by the Deutsche Forschungsgemeinschaft (DFG, German Research Foundation) RTG 2761 LokoAssist (Grant no. 450821862). We also acknowledge support by the DFG within the CRC/TRR 270 (Project-ID 405553726). We are thankful to Lukas Dentler and Ragnar Mogk for being facilitators during the development of this work. We are also grateful to Software Technology Group, Department of Computer Science, and TU Darmstadt for enabling this software development project.